\pdfoutput=1

\documentclass[11pt]{article}

\usepackage{acl}

\usepackage{times}
\usepackage{latexsym}

\usepackage[T1]{fontenc}

\usepackage[utf8]{inputenc}

\usepackage{microtype}

\usepackage{inconsolata}

\usepackage{url}
\usepackage{graphicx}
\usepackage{wrapfig}
\usepackage{amsmath}
\usepackage{amssymb}
\usepackage{color,xcolor,colortbl}
\usepackage{enumitem}
\usepackage{algorithm}
\usepackage{algorithmic}
\usepackage{bm,bbm}
\usepackage{booktabs}
\usepackage{mathtools}
\usepackage{array}
\usepackage{multirow}
\usepackage{soul}
\usepackage{subcaption}
\usepackage{pbox}
\usepackage{pifont}
\usepackage{arydshln}
\usepackage{boldline}
\usepackage{dsfont}
\usepackage{comment}
\usepackage[scaled=0.8]{beramono}

\definecolor{darkblue}{rgb}{0.0, 0.0, 0.55}
\newenvironment{fontpbk}{\fontfamily{lmss}\selectfont}{\par} 
\setul{0.5ex}{0.3ex} 
\setulcolor{blue} 
\usepackage{soul}

\definecolor{d}{HTML}{c6dbef}
\definecolor{l}{HTML}{e5f5e0}
\definecolor{pink1}{RGB}{212, 186, 176}
\definecolor{pink2}{RGB}{127, 134, 123}
\definecolor{pink3}{RGB}{193, 171, 173}
\definecolor{green1}{RGB}{199, 199, 187}
\definecolor{gray1}{RGB}{239, 237, 231}

\title{\emph{MeetingBank}: A Benchmark Dataset for Meeting Summarization}

\author{Yebowen Hu,$^\dagger$ Tim Ganter,$^\ddagger$ Hanieh Deilamsalehy,$^\ddagger$ Franck Dernoncourt,$^\ddagger$\\ 
\textbf{Hassan Foroosh,$^\dagger$ Fei Liu$^\S$}\\[0.8em]
$^\dagger$University of Central Florida \, 
$^\ddagger$Adobe Research \, 
$^\S$Emory University\\
\texttt{huye@knights.ucf.edu \, hassan.foroosh@ucf.edu}\\
\texttt{\quad \{ganter,deilamsa,franck.dernoncourt\}@adobe.com \, fei.liu@emory.edu}
}

\begin{document}
\maketitle

\begin{abstract}

As the number of recorded meetings increases, it becomes increasingly important to utilize summarization technology to create useful summaries of these recordings. However, there is a crucial lack of annotated meeting corpora for developing this technology, as it can be hard to collect meetings, especially when the topics discussed are confidential. Furthermore, meeting summaries written by experienced writers are scarce, making it hard for abstractive summarizers to produce sensible output without a reliable reference. This lack of annotated corpora has hindered the development of meeting summarization technology. In this paper, we present \emph{MeetingBank}, a new benchmark dataset of city council meetings over the past decade. \emph{MeetingBank} is unique among other meeting corpora due to its divide-and-conquer approach, which involves dividing professionally written meeting minutes into shorter passages and aligning them with specific segments of the meeting. This breaks down the process of summarizing a lengthy meeting into smaller, more manageable tasks. The dataset provides a new testbed of various meeting summarization systems and also allows the public to gain insight into how council decisions are made. We make the collection, including meeting video links, transcripts, reference summaries, agenda, and other metadata, publicly available to facilitate the development of better meeting summarization techniques.\footnote{Our dataset can be accessed at: \url{meetingbank.github.io}}

\end{abstract}

\begin{figure*}
\centering
\includegraphics[width=6.2in]{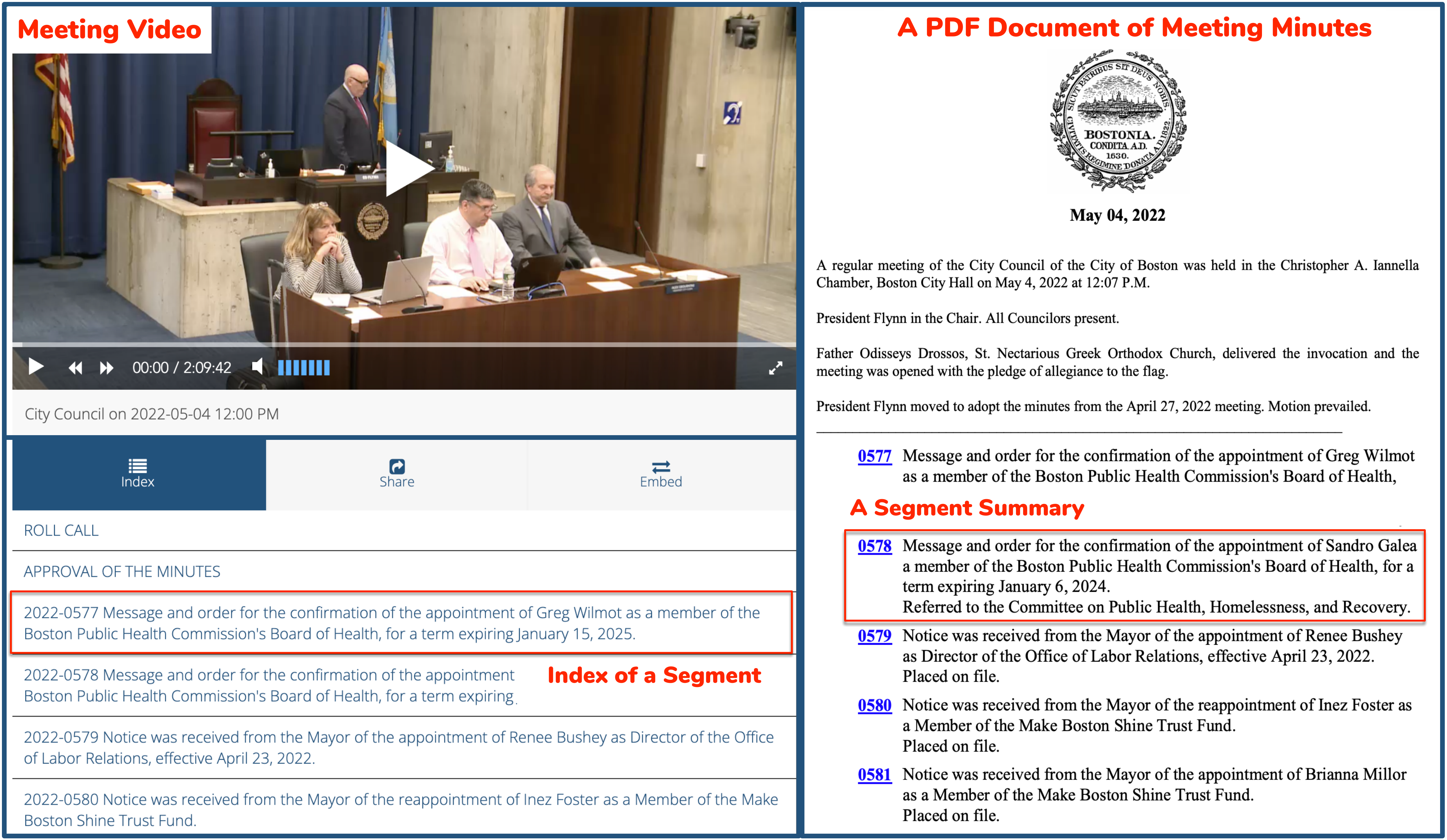}
\caption{A screenshot of a city council meeting of the City of Boston held on May 4, 2022. The meeting video is shown on the left, its corresponding minutes document on the right. The meeting includes discussions of multiple ordinances and resolutions. A summary of the discussion on item 2022-0578 is highlighted in red.}
\label{fig:screenshot_council}
\end{figure*}

\section{Introduction}

An astonishing 55 million meetings happen in the U.S. each week~\cite{Flynn:2022}. With the extensive use of video conferencing software, e.g., Microsoft Teams, Google Meet and Zoom, it has become easier than ever before to record meetings. While these recordings provide a wealth of human intelligence and actionable knowledge, the temporal nature of sound makes it difficult for users to navigate and search for specific content~\cite{Bengio:2004}. A summarization system that produces text summaries from transcripts can help, by providing users with great flexibility in navigating recordings, including but not limited to: meetings, interviews, podcasts, lectures, movies and TV series~\cite{papalampidi-etal-2020-screenplay,zhu-etal-2021-mediasum,song-etal-2022-towards,chen-etal-2022-summscreen,cho-etal-2022-toward}.

Effective meeting summarization requires annotated datasets. Most summarizers, including few-shot and prompt-based~\cite{goyal2022news}, will benefit directly from benchmark datasets containing hundreds of thousands of document-summary pairs such as XSum~\cite{narayan-etal-2018-dont}, MultiNews~\cite{fabbri-etal-2019-multi}, GovReport~\cite{huang-etal-2021-efficient}, PLoS~\cite{goldsack2022}. However, datasets for meeting summarization are relatively scarce, small, or unrepresentative. ICSI and AMI are two benchmark datasets~\cite{Janin:2003,Carletta:2006} that consist of only 75 and 140 meetings, respectively. Other existing datasets for meetings are developed for speech recognition, or are in languages other than English and do not have reference summaries~\cite{tardy-etal-2020-align,kratochvil-etal-2020-large}.

Creating an annotated meeting dataset poses several challenges. First, meetings often contain confidential or proprietary information, making it difficult to share them publicly. Moreover, accurately annotating meeting summaries is a labor-intensive process, even for experienced writers familiar with the meeting topics~\cite{Renals:2007}. Effective meeting summaries should capture key issues discussed, decisions reached, and actions to be taken, while excluding irrelevant discussions~\cite{zechner-2002-automatic,murray-etal-2010-generating}. There thus is a growing need for innovative approaches to construct a meeting dataset with minimal human effort to support advanced meeting solutions.

An increasing number of \emph{city governments} are releasing their meetings publicly to encourage transparency and engage residents in their decision making process. In this paper, we present a systematic approach to develop \textbf{\emph{a city council meeting dataset}}. A city council is the legislative branch of local government. The council members are responsible for making decisions on a range of issues that affect the city and its citizens. These decisions may include creating and approving annual budgets, setting tax rates, confirming appointments of officers, enacting and enforcing ordinances, and setting policies on issues such as land use, public safety, and community development. Figure~\ref{fig:screenshot_council} provides an example of a regular meeting of the City Council of Boston held on May 4, 2022.

We present \textbf{\emph{MeetingBank}}, a benchmark dataset created from the city councils of 6 major U.S. cities to supplement existing datasets. It contains 1,366 meetings with over 3,579 hours of video, as well as transcripts, PDF documents of meeting minutes, agenda, and other metadata. It is an order of magnitude larger than existing meeting datasets~\cite{Carletta:2006}. On average, a council meeting is 2.6 hours long and its transcript contains over 28k tokens, making it a valuable testbed for meeting summarizers and for extracting structure from meeting videos. 
To handle the max sequence length constraint imposed by abstractive summarizers,
we introduce a \emph{divide-and-conquer strategy} to divide lengthy meetings into segments, align these segments with their respective summaries from minutes documents, and keep the segments simple for easy assembly of a meeting summary. This yields 6,892 segment-level summarization instances for training and evaluating of performance. Our repository can be further enhanced through community efforts to add annotations such as keyphrases and queries~\cite{zhong2021qmsum}. To summarize, this paper presents the following contributions:
\begin{itemize}[topsep=5pt,itemsep=0pt,leftmargin=*]

\item We have curated a repository of city council meetings, \emph{MeetingBank}, to advance summarization in an understudied domain. We detail our process of examining 50 major U.S. cities, accessing their city councils' websites for meeting videos and minutes, and obtaining permission to use their data for research purposes. As more cities participate in open government initiatives and release their council meetings, MeetingBank has the potential to continue growing.

\item We test various summarizers including extractive, abstractive with fine-tuning, and GPT-3 with prompting on this task. They are provided with the transcript of a meeting segment and is tasked with generating a concise summary. Experiments with automatic metrics and expert annotators suggest that meeting summarizers should prioritize capturing the main points of meeting discussions and maintaining accuracy to the original.

\end{itemize}

\section{Existing Datasets}
\label{sec:related}

In this section, we review existing meeting datasets, discuss the techniques used to create reference summaries for them and identify research challenges that require attention in this area.

\vspace{0.03in}
\textbf{ICSI} and \textbf{AMI} are widely used datasets for meetings. ICSI~\cite{Janin:2003} is a benchmark of 75 meetings that occurred naturally among speech researchers in a group seminar setting. Each meeting lasts approximately an hour. AMI~\cite{Carletta:2006} contains 100 hours of recorded meetings, including 140 scenario-based meetings where groups of four participants assume roles within a fictitious company to design a product. Meetings typically last around 30-40 minutes, with roles including a project manager, user interface designer, production designer, and marketing expert. A wide range of annotations are performed by annotators, including speech transcriptions, dialogue acts, topics, keywords, extractive and abstractive summaries. Although small in size, these datasets offer a valuable testbed for evaluating meeting summarization systems~\cite{wang-cardie-2013-domain,oya-etal-2014-template,shang-etal-2018-unsupervised,li-etal-2019-keep,koay-etal-2020-domain,koay-etal-2021-sliding,zhang-etal-2022-summn}.

\vspace{0.03in}
Our study complements recent datasets for meetings such as \textbf{{ELITR}} and \textbf{{QMSum}}.
Nedoluzhko et al.~\shortcite{nedoluzhko-etal-2022-elitr} developed ELITR, a dataset of 120 English and 59 Czech technical project meetings spanning 180 hours of content. 
Their minutes are created by participants and specially-trained annotators. Zhong et al.~\shortcite{zhong2021qmsum} developed the QMSum system which extracts relevant utterances from transcripts and then uses the utterances as input for an abstractor to generate query-focused summaries. Human annotators are recruited to collect queries and compose summaries. They annotate a limited number of 25 committee meetings from the Welsh Parliament, 11 from the Parliament of Canada, as well as AMI and ICSI meetings for query-focused summarization.

\begin{figure}[t]
\centering
\includegraphics[width=3in]{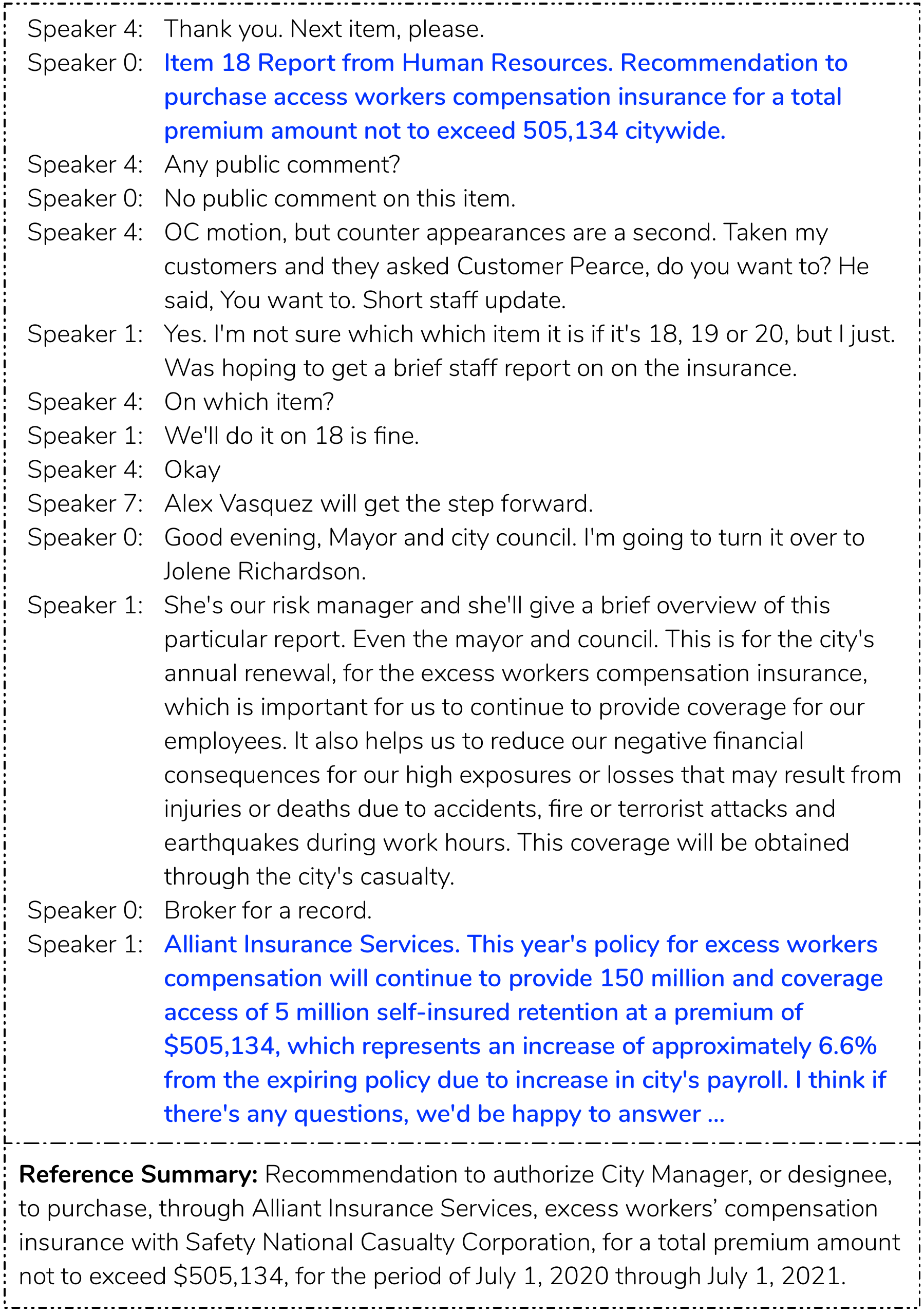}
\caption{An example of a transcript snippet for a meeting segment, which serves as the source text for our summarizer. Similar to BillSum~\cite{kornilova-eidelman-2019-billsum}, a short description of the discussed bill serves as the segment-level reference summary.
Source: Long Beach, 6/23/2022.}
\label{fig:snippet}
\end{figure}

\vspace{0.03in}
Summarization datasets have been developed for genres similar to meetings, such as podcasts, interviews, livestreams and TV series. \textbf{Spotify}~\cite{clifton-etal-2020-100000} released a dataset of 100,000 podcasts to support podcast search and summarization. The dataset includes automatic transcripts with word-level time alignments and creator-provided podcast descriptions are used as reference summaries. \textbf{MediaSum}~\cite{zhu-etal-2021-mediasum} is a dataset of media interviews containing 463.6k transcripts from NPR and CNN, with overview and topic descriptions used as reference summaries. \textbf{StreamHover}~\cite{cho-etal-2021-streamhover} used crowd workers to annotate 370 livestream videos for both extractive and abstractive summaries. \textbf{SummScreen}~\cite{chen-etal-2022-summscreen} consists of TV series transcripts and human-written recaps extracted from community websites. Unlike meetings, these datasets have a smaller number of participants, usually one or two, and do not involve decision making.

\begin{table*}
\setlength{\tabcolsep}{3pt}
\renewcommand{\arraystretch}{1}
\centering
\begin{tabular}{lrrcccrrrrr}
\toprule
& \multicolumn{5}{c}{\textsc{Meeting-Level}} & \multicolumn{1}{c}{\textsc{Inst.}} & \multicolumn{2}{c}{\textsc{Source}} & \multicolumn{2}{c}{\textsc{Summary}}\\
\cmidrule(lr){2-6} \cmidrule(lr){7-7} \cmidrule(lr){8-9} \cmidrule(lr){10-11} 
\multicolumn{1}{l}{\textsc{City}} & \multicolumn{1}{r}{\texttt{\textbf{\#Mtgs}}} & \multicolumn{1}{r}{\texttt{\textbf{\, \#Hrs}}} & {\texttt{\textbf{\, \#Tks \,}}} & \multicolumn{1}{r}{\texttt{\textbf{\#Speakers}}} & \multicolumn{1}{r}{\texttt{\textbf{Period}}} & \multicolumn{1}{r}{\texttt{\textbf{\, \#Segs}}} & \multicolumn{1}{r}{\texttt{\textbf{\, \#Snts}}} & \multicolumn{1}{r}{\texttt{\textbf{\, \#Tks}}} & \multicolumn{1}{c}{\texttt{\textbf{\, \#Snts}}} & \multicolumn{1}{r}{\texttt{\textbf{\#Tks}}}\\
\midrule
Denver & 401 & 979 & 25,460 & [3,20] & 2014--22 & 1,506 & 204 & 5,100 & 3.32 & 111 \\
Seattle & 327 & 446 & 15,045 & [3,14] & 2015--22 & 1,497 & 54 & 1,499 & 1.06 & 78 \\
Long Beach & 310 & 1103 & 39,618 & [4,19] & 2014--22 & 2,695 & 146 & 3,826 & 1.90 & 86\\
Alameda & 164 & 730 & 47,981 & [2,15] & 2015--22 & 672 & 251 & 6,452 & 2.04 & 67\\
King County & 132 & 247 & 20,552 & [2,10] & 2016--22 & 223 & 196 & 5,358 & 1.00 & 78\\
Boston & 32 & 72 & 23,291 & [4,11] & 2021--22 & 299 & 63 & 1,422 & 1.98 & 77\\
\midrule
\multicolumn{11}{l}{\textsc{Total Count}: \textbf{1,366} meetings, \textbf{3,579} hours transcribed, \textbf{6,892} summarization instances collected} \\
\bottomrule
\end{tabular}
\caption{Dataset statistics. Our dataset includes a total of 1,366 city council meetings. We present the number of meetings (\texttt{\#Mtgs}), their cumulative duration in hours (\texttt{\#Hrs}), the average number of tokens per meeting (\texttt{\#Tks}), and the number of speakers per meeting (\texttt{\#Speakers}) for each city. We also provide the number of summarization instances gathered (\texttt{\#Segs}) for each city, as well as the average number of sentences (\texttt{\#Snts}) and tokens (\texttt{\#Tks}) in both source and summary texts. On average, across all cities, a meeting has an average duration of 2.6 hours and 28k tokens. A meeting segment has 2,892 tokens in the source transcript and 87 tokens in the summary.}
\label{tab:stats-per-city}
\end{table*}

\vspace{0.03in}
Meetings can also occur through online chats, as opposed to face-to-face. \textbf{SAMSum}~\cite{gliwa-etal-2019-samsum} is a dataset of 16k chat dialogs with manually annotated abstractive summaries. 
The conversations are short, ranging from 3 to 30 utterances. 
\textbf{ForumSum}~\cite{khalman-etal-2021-forumsum-multi} is another dataset that includes online posts collected from internet forums, with associated human-written summaries. 
\textbf{DialogSum}~\cite{chen-etal-2021-dialogsum} contains 13k dialogs gathered from multiple websites.
Medical consultations conducted through online chats have also been used to create consultation summaries~\cite{zeng-etal-2020-meddialog,laleye-etal-2020-french,moramarco-etal-2021-preliminary,gao-wan-2022-dialsummeval}. Our paper focuses on developing a dataset of naturally-occurring meeting conversations to aid in the development of summarization systems from transcripts.

\section{Creation of \textbf{\emph{MeetingBank}}}
\label{sec:data}

There is a growing need to make public meetings more accessible and inclusive for citizens to engage with their local officials and have their voices heard. A 2020 report from the American Academy of Arts and Sciences\footnote{\url{amacad.org/ourcommonpurpose/recommendations}} reveals that only 11\% of Americans attend public meetings to discuss local issues. Organizations such as the \href{https://councildataproject.org/}{CouncilDataProject.org} and \href{https://codeforamerica.org/}{CodeforAmerica.org} are working to improve the search infrastructure of public meetings. Creation of a city council meeting dataset could provide a valuable testbed for meeting summarizers, and an effective summarizer could make public meetings more accessible by allowing citizens to navigate meetings more efficiently, thus promoting citizen engagement.

We begin by compiling a list of the top 50 cities in the U.S. by population\footnote{\url{www.infoplease.com/us/cities/top-50-cities-us-population-and-rank}} and narrow it down to include only cities that regularly release meetings with accompanying minutes documents, and have downloadable videos on their city council websites. An example of a public meeting and its minutes can be seen in Figure~\ref{fig:screenshot_council}. We consult with our legal team and reach out to city councils when necessary to ensure compliance with licensing and data policies.\footnote{For example, we have excluded the City of San Francisco from our dataset as the city has advised us that meeting videos may be reposted or edited with attribution, but minutes and agenda are official public documents that are not permitted to be reposted or edited.} Our dataset for this release includes \textbf{1,366 meetings} from six cities or municipalities spanning over a decade, including \emph{Seattle, Washington; King County, Washington; Denver, Colorado; Boston, Massachusetts; Alameda, California; and Long Beach, California}.

Using minutes documents as-is for the development of summarization systems can be challenging. This is because minutes are often provided in PDF format and do not always align with the flow of meeting discussions. For instance, minutes may include a section on the Mayor's update that provides detailed information on appointments of officers, but this is only briefly mentioned in the meeting. In general, \emph{minutes} are more formal and comprehensive records of meetings, including information such as the date, location, attendees, summary of main points discussed, decisions made, and action items assigned. Minutes are distributed to the stakeholders after the meeting. In contrast, \emph{meeting summaries} tend to be shorter and less formal, focusing on the key points discussed in a meeting.

We propose a divide-and-conquer strategy for creating reference meeting summaries. It involves dividing lengthy meetings into segments, aligning them with their corresponding summaries from minutes documents, and keeping the segments simple for easy assembly of a meeting summary. To start, we extract a list of Council Bill (CB) numbers discussed at the meeting by parsing the minutes and city council websites.\footnote{E.g., \url{boston.legistar.com/MeetingDetail.aspx?ID=958849&GUID=CAD14B15-407D-4552-AF01-4BD64314AD2D}} For each bill, we then identify a short description that summarizes its content, which serves as the reference summary. Next, we use the bill number to obtain the corresponding meeting segment, including its start and end time, by referencing the index of the meeting on the city council website (see Figure~\ref{fig:screenshot_council}). The transcript of that segment serves as the source text for the summarizer. After filtering out noisy and too short segments,\footnote{We require a minimum length of 60 seconds for a segment to be included in our dataset, as segments shorter than this are too brief to be summarized. The reference summary for a segment should contain at least 10 words.} we have a total of \textbf{6,892 segment-level instances} in our dataset (Table~\ref{tab:stats-per-city}).

We use \href{Speechmatics.com}{Speechmatics.com}'s speech-to-text API to automatically transcribe 3,579 hours of meetings, an order of magnitude larger than existing datasets. Our transcripts include word-level time alignment, casing, punctuation, and speaker diarization. City council meetings range from 2 to 19 speakers, with an average duration of 2.6 hours and 28,358 tokens per transcript. On average, across all cities, a meeting segment has 2,892 tokens in the transcript and 87 tokens in the summary. The resulting compression rate is 97\%. For every council meeting, we collect the following information, represented using their attribute name and sample value:\footnote{We gather the meeting agenda and other supporting documents when available. Our focus is on summarizing transcripts of spontaneous speech where natural language is the primary means of information conveyance. We do not attempt to obtain non-verbal cues such as eye gazes, facial expressions, laughter, or understand persuasive argumentation.}

\vspace{0.05in}
\begin{enumerate}[topsep=3pt,itemsep=-2pt,leftmargin=*]
\begin{small}   
\begin{fontpbk}
\item Title of the meeting (``\emph{Full Council 12/14/15}'')
\item Meeting ID (``\emph{SeattleCityCouncil\_12142015}'')
\item Link to the specific meeting\\
(\emph{\url{https://www.seattlechannel.org/FullCouncil?videoid=x60447&Mode2=Video}})
\item Link to the meeting video\\
(\emph{\url{https://video.seattle.gov/media/council/full_121415V.mp4}})
\item Link to the meeting minutes \\
(\url{https://seattle.legistar.com/View.ashx?M=M&ID=449835&GUID=712D0B7C-A536-498E-8C99-D3037AE814D9})
train\item ID of a specific topic discussed (``\emph{CB 118549}'')
\item Type of the ID number (``\emph{Ordinance}'')
\item Start and end times in the video where the topic was discussed (``\emph{00:06:24}'' to ``\emph{00:18:19}'')
\item Full transcript of the video, along with start and end points of each segment of the meeting (Figure~\ref{fig:snippet})
\item Reference summary for each meeting segment
\end{fontpbk}
\end{small}
\end{enumerate}

\begin{table*}
\setlength{\tabcolsep}{3.5pt}
\renewcommand{\arraystretch}{1}
\centering
\begin{tabular}{lcccccccccr}
\toprule
& \multicolumn{4}{c}{\textsc{ROUGE}} & \multicolumn{2}{c}{\textsc{BLEU + MET.}} & \multicolumn{2}{c}{\textsc{Embeddings}} & \multicolumn{1}{c}{\textsc{QA}} & \multicolumn{1}{c}{\textsc{Summ}} \\
\cmidrule(lr){2-5} \cmidrule(lr){6-7} \cmidrule(lr){8-9} \cmidrule(lr){10-10} 
\textsc{Model} & \texttt{\textbf{\, R-1 \,}} & \texttt{\textbf{\, R-2 \,}} & \texttt{\textbf{\, R-L \,}} & \texttt{\textbf{R-We}} & \texttt{\textbf{\, BLEU \,}} & \texttt{\textbf{METEOR}} & \texttt{\textbf{BERTS.}} & \texttt{\textbf{MoverS.}} & \texttt{\textbf{QAEval}} & \textsc{Len.}\\
\midrule
Extr Oracle & 61.82 & {46.60} & 52.61 & 55.60 & 22.99 & 52.35 & 69.54 & 63.15 & 21.69 & 64.89\\					
Lead-3 & 28.15 & {19.53} & 25.75 & 23.77 & 7.90 & 23.53 & 50.20 & 54.56 & 9.62 & 40.79\\	
LexRank & 24.61 & 10.68 & 19.06 & 15.98 & 5.86 & 17.70 & 48.55 & 53.23 & 6.53 & 53.70\\
TexRank & 30.25 & 15.97 & 24.37 & 21.91 & 9.16 & 22.10 & 52.32 & 54.65 & 8.33 & 61.81\\
\midrule
BART w/o FT & 31.02 & 16.76 & 23.93 & 23.11 & 8.07 & 16.63 & 53.04 & 53.91 & 13.63 & 140.65\\	
HMNet & 50.55 & 34.22 & 45.07 & 45.05 & 13.93 & 46.80 & 66.38 & 60.34 & 12.96 & 50.44\\
Longformer & 59.89 & 48.23 & 55.66 & 56.15 & 40.04 & 50.92 & 75.31 & 65.27 & 23.54 & 82.86\\
BART & 62.81 & 51.66 & 58.84 & 59.32 & 41.46 & 53.24 & 77.17 & 66.74 & 26.87 & 89.46\\
Pegasus & 68.54 & 59.28 & 66.09 & 65.75 & 33.29 & \textbf{70.24} & 80.70 & 70.44 & \textbf{27.13} & 49.90\\
DialogLM & \textbf{70.30} & \textbf{60.12} & \textbf{67.54} & \textbf{67.55} & \textbf{45.42} & {66.44} & \textbf{81.61} & \textbf{71.56} & 25.75 & 66.36\\
\midrule
GPT3-D3 & 36.37 & 16.95 & 26.82 & 26.14 & 8.80 & 25.41 & 56.53 & 55.61 & 10.88 & 60.41\\
\bottomrule
\end{tabular}
\caption{Evaluation of state-of-the-art summarization systems on the test split of our city council dataset.
The final column shows the average length of system summaries, which are generated by each individual summarizer using their default settings.
}
\label{tab:results-test-set}
\end{table*}

\section{Data Analysis}
\label{sec:analysis}

\begin{figure}[t]
\includegraphics[width=3in]{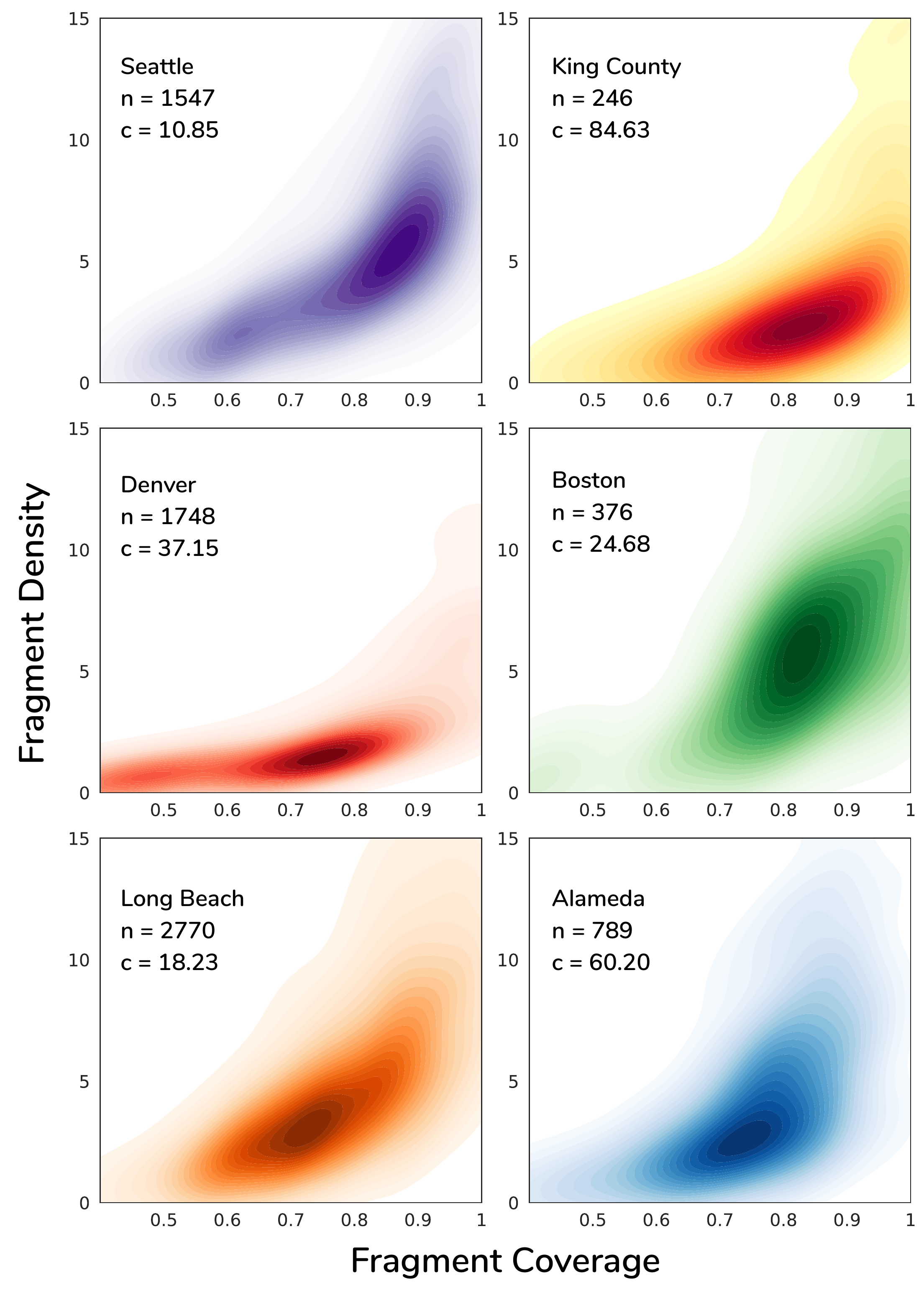}
\vspace{-0.05in}
\caption{$\mathcal{C}$\emph{overage} and $\mathcal{D}$\emph{ensity} scores for segment-level summarization instances, plotted for individual cities. Seattle and Boston have the highest density scores among the cities studied, while Denver has the lowest, indicating that the minutes for this city have undergone a high degree of editing.}
\label{fig:coverage-density}
\vspace{-0.15in}
\end{figure}

We measure the level of abstraction in meeting summaries by quantifying the amount of reused text. Higher abstraction poses more challenges to the meeting summarization systems. We employ two common measures, $\mathcal{C}$\emph{overage} and $\mathcal{D}$\emph{ensity}~\cite{grusky-etal-2018-newsroom}, to evaluate segment-level reference summaries. Results are illustrated in Figure~\ref{fig:coverage-density}, with coverage on x-axis and density on y-axis.

The $\mathcal{C}$\emph{overage} score measures the percentage of summary words that appear in the source transcript. E.g., a summary of 10 words that includes 8 words from the source transcript and 2 new words has a coverage score of 0.8 (= 8/10). As shown in the figure, the $\mathcal{C}$\emph{overage} score for city council meeting summaries is in the range of 0.7-0.9 for most cities. This suggests that, unlike news, these meeting summaries tend to include discussion points verbatim rather than performing abstraction. Given a compression rate of over 90\%, an effective summarizer should focus on accurately identifying content to be included in the summary.

A $\mathcal{D}$\emph{ensity} score evaluates how much a summary can be characterized as a set of extractive fragments. For example, a summary of 10 words made up of two extractive fragments of length 1 and 6 and three new words would have a density score of 3.7 = (1$^2$ + 6$^2$)/10. A summary with long consecutive text fragments taken from the source transcript would yield a high density score. We observe that Seattle and Boston have the highest density scores among all cities studied, while Denver has the lowest score, indicating a high degree of editing is performed to produce the  minutes for Denver. We note that certain resolutions and ordinances are read out plainly at the council meetings and included in the minutes, making the summaries often have higher density scores than those of news documents.

The $\mathcal{C}$\emph{overage} and $\mathcal{D}$\emph{ensity} measures can be influenced by a range of factors such as the length and complexity of meetings and the preferences of the minute-takers. The diversity of meeting summaries highlights the complexity of this task.

\section{Performance of Existing Systems}
\label{sec:baselines}

We evaluate state-of-the-art summarization systems on city council meetings, focusing on segments of the meetings rather than entire transcripts due to the length constraint imposed by abstractive summarizers. We split our dataset into train, validation and test sets, containing 5169, 861, 862 instances respectively. Each summarizer is given the transcript of a meeting segment and tasked with generating a concise summary. The results are reported for the test set of our meeting dataset.

\vspace{0.03in}
\noindent\textbf{\emph{Extractive.}}\quad\, Our extractive methods include the \texttt{Oracle}, \texttt{LEAD}, \texttt{LexRank} and \texttt{TextRank}~\cite{Erkan:2004,mihalcea-tarau-2004-textrank}. The \texttt{Extractive Oracle}\footnote{\url{github.com/pltrdy/extoracle_summarization}} selects the highest-scoring sentences from the input transcript, adding one sentence at a time until the combined R1 and R2 score can no longer be improved. The \texttt{LEAD-N} baseline selects the first N sentences of the input. \texttt{LexRank} and \texttt{TextRank}, both graph-based methods, determine the importance of sentences by analyzing their centrality in the graph structure. Both methods are set to extract two sentences from a transcript segment, which is the average number of sentences in the reference summaries.

\begin{table*}
\setlength{\tabcolsep}{3.9pt}
\renewcommand{\arraystretch}{1}
\centering
\begin{tabular}{lcccc|rrrrrr}
\toprule
& & \multicolumn{3}{c}{\textsc{Test Set (All)}} & \multicolumn{6}{c}{\textsc{Test Set (By City)}} \\
\cmidrule(lr){3-5} \cmidrule(lr){6-11} 
\multicolumn{1}{l}{\textsc{Train}} & \multicolumn{1}{r}{\texttt{\textbf{\#Inst.}}} & \multicolumn{1}{r}{\texttt{\textbf{\, R-1 \,}}} & \multicolumn{1}{r}{\texttt{\textbf{\, R-2 \,}}} & \multicolumn{1}{r}{\texttt{\textbf{\, R-L \,}}} & \multicolumn{1}{r}{\texttt{\textbf{L.B.}}}\, & \multicolumn{1}{r}{\texttt{\textbf{Denver}}} & \multicolumn{1}{r}{\texttt{\textbf{Seattle}}} & \multicolumn{1}{c}{\texttt{\textbf{Alameda}}} & \multicolumn{1}{r}{\texttt{\textbf{Boston}}} & \multicolumn{1}{r}{\texttt{\textbf{K.C.}}}\\
\cmidrule(lr){1-1} \cmidrule(lr){2-2} \cmidrule(lr){3-5} \cmidrule(lr){6-11} 
w/o L.B. & 3,157 & 60.36 & 48.57 & 56.48 & \textbf{21.30}$\downarrow$ & 1.95$\uparrow$ & 4.39$\uparrow$ & 1.72$\downarrow$ & 3.92$\downarrow$ & 3.71$\downarrow$\\
w/o Denver & 3,951 & 58.74 & 46.90 & 54.31 & 2.84$\downarrow$ & \textbf{32.43}$\downarrow$ & 1.96$\downarrow$ & 0.22$\downarrow$ & 0.22$\downarrow$ & 1.64$\uparrow$\\
w/o Seattle & 4,115 & \textbf{53.42} & \textbf{40.05} & \textbf{48.28} & 3.62$\downarrow$ & 3.34$\uparrow$ & \textbf{31.12}$\downarrow$ & 2.19$\downarrow$ & 1.06$\downarrow$ & 2.79$\downarrow$\\
w/o Alameda & 4,698 & 61.54 & 50.31 & 57.48 & 2.41$\downarrow$ & 4.07$\uparrow$ & 3.00$\uparrow$ & \textbf{19.32}$\downarrow$ & 2.54$\downarrow$ & 1.68$\downarrow$\\
w/o Boston & 4,936 & 62.70 & 51.62 & 58.82 & 1.21$\uparrow$ & 0.75$\uparrow$ & 7.52$\downarrow$ & 1.02$\uparrow$ & \textbf{42.82}$\downarrow$ & 4.96$\uparrow$\\
w/o K.C. & 4,988 & 63.60 & 52.71 & 59.87 & 3.60$\downarrow$ & 3.32$\uparrow$ & 4.37$\uparrow$ & 6.14$\downarrow$ & 3.76$\downarrow$ & \textbf{11.6}$\downarrow$\\
\bottomrule
\end{tabular}
\caption{Evaluation of the BART summarizer using a series of ablations. \textsc{Left:} we remove all the training instances from a single city and fine-tune the model with the remaining instances, denoted by \texttt{\#Inst}. We find that although the City of Seattle only contributes a moderate number of training instances, removing them has led to a substantial decrease in summarization performance. \textsc{Right:} we evaluate the performance of the BART summarizer on a city-by-city basis. We show the variance in R-2 F-scores for each test city when training instances from the same city are included vs. when they are excluded. $\downarrow$ indicates a performance drop and $\uparrow$ a performance gain.}
\label{tab:results-per-city}
\end{table*}

\vspace{0.03in}
\noindent\textbf{\emph{Abstractive with fine-tuning.}}\quad
We investigate five best-performing neural abstractive summarizers. These include \texttt{BART}-large~\cite{lewis-etal-2020-bart}, a denoising autoencoder that is trained to reconstruct original text from corrupted input, \texttt{Pegasus}~\cite{zhang2020pegasus}, a model that is trained to regenerate missing key sentences, \texttt{Longformer}~\cite{beltagy2020longformer}, a model designed to handle long sequences through windowed attention, \texttt{DialogLM}~\cite{zhong2022dialoglm}, a summarizer developed for summarizing long dialogues and pretrained using window-based denoising and \texttt{HMNet}~\cite{zhu-etal-2020-hierarchical}, a hierarchical model that uses role vectors to distinguish between speakers. We evaluate \texttt{BART}-large with and without fine-tuning on our dataset, and compare the results to other models.

\vspace{0.03in}
\noindent\textbf{\emph{GPT-3 with prompting.}}\quad
Large language models like GPT-3, T5 and PaLM~\cite{brown2020language,raffel2020exploring,chowdhery2022palm} have developed advanced capabilities due to increased computation, parameters, training data size~\cite{wei2022emergent}. When prompted, GPT-3 can generate a summary of a source text by identifying the most important information. The \texttt{text-davinci-003} version of GPT-3 is used in this study, with a prompt asking the model to summarize the text in two sentences~\cite{goyal2022news}.

\vspace{0.03in}
\noindent\textbf{\emph{Evaluation Metrics.}}\,\,
We use a variety of automatic evaluation metrics to assess the quality of transcript summaries. These metrics are broadly grouped into three categories:
(a) traditional metrics comparing system and reference summaries based on lexical overlap, including ROUGE~\citep{lin-2004-rouge}, ROUGE-we (w/ embeddings), BLEU~\citep{post-2018-call} and METEOR~\citep{banerjee-lavie-2005-meteor};
(b) new metrics making use of contextualized embeddings to measure semantic similarity, e.g., BertScore~\citep{Zhang:2020:BERTScore} and MoverScore~\citep{zhao-etal-2019-moverscore};
(c) question answering-based metrics, where the hypothesis is that high-quality summaries should contain informative content and act as a surrogate for the original document in satisfying users' information needs. We leverage summarization evaluation toolkits provided by Fabbri et al.\shortcite{Fabbri:2021}, SacreBLEU~\cite{post-2018-call}, QAEval~\cite{deutsch-etal-2021-towards} and SummerTime~\cite{ni-etal-2021-summertime} to report results using these metrics.

Table~\ref{tab:results-test-set} shows our experimental results. We observe that the Extractive Oracle yields a high R-2 F-score of 46.60\%, indicating that the content of reference summaries mostly comes from the source transcripts, and extractive summarization methods could be promising. However, it would be desirable to develop more sophisticated methods than LexRank and TextRank, despite their outstanding performance on news articles, they do not perform well on this task. We find that DialogLM performs the best among abstractive summarizers. This is not surprising as it is designed for summarizing long dialogues. Pegasus also demonstrates strong performance, its results are on par with those of DialogLM. Fine-tuning BART on in-domain data yields substantial improvement on its performance. Finally, GPT-3 with prompting does not perform well according to automatic metrics, but we have interesting findings during human assessment (\S\ref{sec:human-eval}).\footnote{We find that some automatic metrics are affected by the extractiveness of summaries, such as MoverScore and ROUGE, while others, such as BERTScore and QAEval, are less sensitive. Metrics that are sensitive to extractiveness give varying scores across different datasets, and those that are insensitive tend to produce scores in similar ranges.}

\begin{table*}
\setlength{\tabcolsep}{5pt}
\renewcommand{\arraystretch}{1.2}
\centering
\textsf{
\begin{small}
\begin{tabular}{ll}
\rowcolor{pink1!50} \textbf{\emph{Informativeness}}: & How well does the summary capture the main points of the meeting segment? A good \\
\rowcolor{pink1!50} & summary should contain all and only the important information of the source.\\[0.3em]
\rowcolor{gray1!50} \textbf{\emph{Factuality}}: & Are the facts provided by the summary consistent with facts in the meeting segment? A \\
\rowcolor{gray1!50} & good summary should reproduce all facts accurately and not make up untrue information.\\[0.3em]
\rowcolor{pink1!50} \textbf{\emph{Fluency}}: & Consider the individual sentences of the summary, are they well-written and grammaticall?\\[0.3em]
\rowcolor{gray1!50} \textbf{\emph{Coherence}}: & Consider the summary as a whole, does the content fit together and sound natural? A \\
\rowcolor{gray1!50} & good summary should not just be a collection of related information, but should build from \\
\rowcolor{gray1!50} & sentence to sentence to a coherent body of information about a topic.\\[0.3em]
\rowcolor{pink1!50} \textbf{\emph{Redundancy}}: & Does the summary contain redundant content? A good summary should not have \\
\rowcolor{pink1!50} & unnecessary word or phrase repetitions in a sentence or semantically similar sentences.\\
\end{tabular}
\end{small}}
\caption{Human evaluation criteria, adapted from Fabbri et al.~\shortcite{Fabbri:2021}.}
\label{tab:eval-criteria}
\end{table*}

\section{City-by-City Analysis}
\label{sec:ablations}

We investigate the characteristics that make effective training instances for meeting summarization by conducting a series of ablations. We begin by removing all training instances from a single city and using the remaining instances to fine-tune the BART summarizer. The results are shown in Table~\ref{tab:results-per-city} (left panel), where we present the R-1, R-2, and R-L F-scores. We find that although the City of Seattle only contributes a moderate number of training instances, removing them has led to a substantial decrease in summarization performance, resulting in an R-2 F-score of 40.05\%. It suggests that these training instances are quite effective and the City Council of Seattle might have implemented a better practice of producing high-quality meeting minutes compared to other cities.

\begin{table*}
\setlength{\tabcolsep}{3.3pt}
\renewcommand{\arraystretch}{1}
\centering
\begin{tabular}{lccccccc}
\toprule
& \multicolumn{2}{c}{\textsc{Extractive}} & \multicolumn{4}{c}{\textsc{Abstractive w/ Finetuning}} & \textsc{Prompting}\\
\cmidrule(lr){2-3} \cmidrule(lr){4-7} \cmidrule(lr){8-8} 
\textsc{Criterion} & \texttt{\textbf{LEAD}} & \texttt{\textbf{LexRank}} & \texttt{\textbf{HMNet}} & \texttt{\textbf{BART}} & \texttt{\textbf{DialogLM}} & \texttt{\textbf{Pegasus}} & \texttt{\textbf{GPT-3}}\\
\midrule
Informativeness & 1.72{\small $\pm$1.22} & 1.90{\small $\pm$1.15} & 2.44{\small $\pm$1.15} & 3.43{\small $\pm$1.13} & 3.50{\small $\pm$1.15} & 3.65{\small $\pm$1.10} & \textbf{3.74}{\small $\pm$1.17} \\
Factuality & 1.92{\small $\pm$1.35} & 2.42{\small $\pm$1.31} & 2.65{\small $\pm$1.18} & 3.45{\small $\pm$1.19} & 3.58{\small $\pm$1.09} & 3.79{\small $\pm$1.13} & \textbf{3.82}{\small $\pm$1.09} \\
Fluency & 2.89{\small $\pm$1.50} & 3.22{\small $\pm$1.32} & 2.78{\small $\pm$1.32} & 3.58{\small $\pm$1.13} & 3.47{\small $\pm$1.14} & 3.71{\small $\pm$1.23} & \textbf{4.52}{\small $\pm$0.83}\\
Coherence & 2.14{\small $\pm$1.34} & 2.70{\small $\pm$1.39} & 2.74{\small $\pm$1.39} & 3.64{\small $\pm$1.15} & 3.72{\small $\pm$1.13} & 3.88{\small $\pm$1.13} & \textbf{4.41}{\small $\pm$1.00}\\
Redundancy & 3.79{\small $\pm$1.38} & 3.53{\small $\pm$1.40} & 3.42{\small $\pm$1.40} & 3.54{\small $\pm$1.41} & 3.96{\small $\pm$1.35} & 4.15{\small $\pm$1.25} & \textbf{4.57}{\small $\pm$0.75}\\
\midrule
\textsc{Average Score} & 2.49 & 2.75 & 2.81 & 3.53 & 3.65 & {3.84} & \textbf{4.21}\\
\bottomrule
\end{tabular}
\caption{Human evaluation results. We observe that abstractive systems perform stronger than extractive systems. GPT-3 is well received in human assessments, but still falls short in terms of informativeness and factuality.}
\label{tab:human-eval}
\end{table*}

We evaluate the performance of the BART summarizer on a city-by-city basis. 
We show the variance in R-2 F-scores for each test city when training instances from the same city are included vesus when they are excluded, as seen in the right panel of Table~\ref{tab:results-per-city}. For instance, we observe a performance drop ($\downarrow$) of 32.43\% for the City of Denver when all training instances from the same city are removed from fine-tuning.\footnote{We fine-tune the BART summarizers for the same number of steps in both cases to mitigate the impact of varying number of training instances.} 
We observe that Seattle, Boston, and Denver benefit more from fine-turning using same-city training data. Particularly, Seattle and Boston have shorter source transcripts and their reference summaries tend to reuse texts from the source. It suggests that different cities may have varying levels of discussions in council meetings and different styles of meeting minutes, and that training instances from the same city are crucial for achieving the best performance.

\section{Human Evaluation}
\label{sec:human-eval}

We evaluate the performance of seven state-of-the-art summarization systems, including fine-tuned abstractive models \texttt{HMNet}, \texttt{BART}, \texttt{Pegasus}, \texttt{DialogLM}, \texttt{GPT-3} with prompting, and traditional extractive models \texttt{LexRank} and \texttt{LEAD} to best assess the effectiveness of system-generated meeting summaries. All abstractive models have been fine-tuned on the train split of our city councils dataset to achieve the best possible results.

To ensure high quality in the assessment of summaries, we have worked with \url{iMerit.net}, a labor sourcing company, to recruit experienced evaluators from the U.S. and India to perform annotations. The workers are registered on \url{Appen.com}, a crowdsourcing platform, to complete the tasks and deliver results. A total of three workers from the United States and six workers from India participate in our evaluations, including pilot annotations.\footnote{All workers have excellent English proficiency, with U.S. workers being native speakers. After a pilot annotation, we decide to work with only U.S. workers due to their high quality of work. They are compensated at \$27.50/hr.}

The workers are asked to watch a video segment, typically 30 minutes or less, read the transcript, and then evaluate the quality of each system summary based on five criteria: \emph{informativeness}, \emph{factuality}, \emph{fluency}, \emph{coherence}, and \emph{redundancy}. These criteria are outlined in Table~\ref{tab:eval-criteria}. 
Importantly, summaries are presented in a random order to prevent workers from making assumptions about quality based on the order they are presented.

In Table~\ref{tab:human-eval}, we present the performance of summarization systems on 200 randomly selected instances. A 5-point Likert scale is used to evaluate each criterion. The scores are then averaged, and standard deviation is also reported. We find that among the five criteria, redundancy is the least of concern. Furthermore, we observe that abstractive systems perform stronger than extractive systems. The best-performing abstractive system is \texttt{Pegasus}. We believe its effectiveness is attributed to the pre-training method of masking key sentences within a document and using the remaining sentences to regenerate them, making it particularly well-suited for this task and effective at identifying important content from the transcripts.

We find that \texttt{GPT-3} achieves the highest overall score of 4.21 according to human evaluations across all criteria. 
This aligns with recent studies that demonstrate GPT-3's near-human performance in news summarization~\cite{goyal2022news,zhang2023benchmarking}.
On our meeting dataset, \texttt{GPT-3} shows exceptional performance in terms of fluency and coherence, receiving scores of 4.52 and 4.41 respectively. However, its results are less impressive in terms of informativeness and factuality with scores of 3.74 and 3.82, but still on par with the best abstractive model, \texttt{Pegasus}. Our findings suggest that meeting summarization solutions should continue to focus on capturing the main discussion points and staying true to the original content.

\section{Conclusion}
\label{sec:discussion}

We created a benchmark dataset from city council meetings and tested various summarization systems including extractive, abstractive with fine-tuning, and GPT-3 with prompting on this task. Our findings indicate that GPT-3 is well received in human assessments, but it falls short in terms of informativeness and factual consistency. Our MeetingBank dataset could be a valuable testbed for researchers designing advanced meeting summarizers and for extracting structure from meeting videos.

\section{Limitations}

We present a new dataset for meeting summarization that has the potential to improve the efficiency and effectiveness of meetings. However, we note that the dataset is limited to city council meetings from U.S. cities over the past decade and licensing issues have restricted our ability to include certain city council meetings in the dataset. For example, we contacted the City Council of San Francisco and were informed that they do not allow the redistribution of meeting minutes. Moreover, our dataset does not include non-verbal cues such as eye gazes, gestures and facial expressions, which may make it less suitable for developing summarization systems that rely on these cues. Despite these limitations, we believe that the dataset is of high quality and will be a valuable resource for the development of meeting summarization systems.

\section{Ethical Considerations}

The city council meetings included in this dataset are publicly accessible. We obtain meeting videos, minutes documents, and other metadata from publicly available sources. We consult with our legal team and reach out to city councils as necessary to ensure compliance with licensing and data policies. We release this dataset to facilitate the development of meeting summarization systems and have made efforts to ensure that the dataset does not include confidential information. Our dataset is intended for research purposes only.

\section*{Acknowledgements}
We are grateful to the reviewers for their insightful feedback, which has helped enhance the quality of our paper. This research has been partially supported by the NSF CAREER award, \#2303655.

\bibliography{anthology,custom,trans_summ}

\begin{thebibliography}{59}
\expandafter\ifx\csname natexlab\endcsname\relax\def\natexlab#1{#1}\fi

\bibitem[{Banerjee and Lavie(2005)}]{banerjee-lavie-2005-meteor}
Satanjeev Banerjee and Alon Lavie. 2005.
\newblock \href {https://aclanthology.org/W05-0909} {{METEOR}: An automatic
  metric for {MT} evaluation with improved correlation with human judgments}.
\newblock In \emph{Proceedings of the {ACL} Workshop on Intrinsic and Extrinsic
  Evaluation Measures for Machine Translation and/or Summarization}, pages
  65--72, Ann Arbor, Michigan. Association for Computational Linguistics.

\bibitem[{Beltagy et~al.(2020)Beltagy, Peters, and
  Cohan}]{beltagy2020longformer}
Iz~Beltagy, Matthew~E Peters, and Arman Cohan. 2020.
\newblock Longformer: The long-document transformer.
\newblock \emph{arXiv preprint arXiv:2004.05150}.

\bibitem[{Bengio and Bourlard(2004)}]{Bengio:2004}
Samy Bengio and Hervé Bourlard. 2004.
\newblock \emph{Machine Learning for Multimodal Interaction: First
  International Workshop}.
\newblock Springer Berlin, Heidelberg.

\bibitem[{Brown et~al.(2020)Brown, Mann, Ryder, Subbiah, Kaplan, Dhariwal,
  Neelakantan, Shyam, Sastry, and Askell~et al}]{brown2020language}
Tom Brown, Benjamin Mann, Nick Ryder, Melanie Subbiah, Jared~D Kaplan, Prafulla
  Dhariwal, Arvind Neelakantan, Pranav Shyam, Girish Sastry, and Amanda
  Askell~et al. 2020.
\newblock \href
  {https://proceedings.neurips.cc/paper/2020/file/1457c0d6bfcb4967418bfb8ac142f64a-Paper.pdf}
  {Language models are few-shot learners}.
\newblock In \emph{Advances in Neural Information Processing Systems},
  volume~33, pages 1877--1901.

\bibitem[{Carletta et~al.(2006)Carletta, Ashby, Bourban, Flynn, Guillemot,
  Hain, Kadlec, and Karaiskos~et al.}]{Carletta:2006}
Jean Carletta, Simone Ashby, Sebastien Bourban, Mike Flynn, Mael Guillemot,
  Thomas Hain, Jaroslav Kadlec, and Vasilis Karaiskos~et al. 2006.
\newblock The {AMI} meeting corpus: A pre-announcement.
\newblock In \emph{Machine Learning for Multimodal Interaction}, pages 28--39.
  Springer Berlin Heidelberg.

\bibitem[{Chen et~al.(2022)Chen, Chu, Wiseman, and
  Gimpel}]{chen-etal-2022-summscreen}
Mingda Chen, Zewei Chu, Sam Wiseman, and Kevin Gimpel. 2022.
\newblock \href {https://doi.org/10.18653/v1/2022.acl-long.589}
  {{S}umm{S}creen: A dataset for abstractive screenplay summarization}.
\newblock In \emph{Proceedings of the 60th Annual Meeting of the Association
  for Computational Linguistics (Volume 1: Long Papers)}, pages 8602--8615,
  Dublin, Ireland. Association for Computational Linguistics.

\bibitem[{Chen et~al.(2021)Chen, Liu, Chen, and
  Zhang}]{chen-etal-2021-dialogsum}
Yulong Chen, Yang Liu, Liang Chen, and Yue Zhang. 2021.
\newblock \href {https://doi.org/10.18653/v1/2021.findings-acl.449}
  {{D}ialog{S}um: {A} real-life scenario dialogue summarization dataset}.
\newblock In \emph{Findings of the Association for Computational Linguistics:
  ACL-IJCNLP 2021}, pages 5062--5074, Online. Association for Computational
  Linguistics.

\bibitem[{Cho et~al.(2021)Cho, Dernoncourt, Ganter, Bui, Lipka, Chang, Jin,
  Brandt, Foroosh, and Liu}]{cho-etal-2021-streamhover}
Sangwoo Cho, Franck Dernoncourt, Tim Ganter, Trung Bui, Nedim Lipka, Walter
  Chang, Hailin Jin, Jonathan Brandt, Hassan Foroosh, and Fei Liu. 2021.
\newblock \href {https://doi.org/10.18653/v1/2021.emnlp-main.520}
  {{S}tream{H}over: Livestream transcript summarization and annotation}.
\newblock In \emph{Proceedings of the 2021 Conference on Empirical Methods in
  Natural Language Processing}, pages 6457--6474, Online and Punta Cana,
  Dominican Republic. Association for Computational Linguistics.

\bibitem[{Cho et~al.(2022)Cho, Song, Wang, Liu, and Yu}]{cho-etal-2022-toward}
Sangwoo Cho, Kaiqiang Song, Xiaoyang Wang, Fei Liu, and Dong Yu. 2022.
\newblock \href {https://aclanthology.org/2022.emnlp-main.8} {Toward unifying
  text segmentation and long document summarization}.
\newblock In \emph{Proceedings of the 2022 Conference on Empirical Methods in
  Natural Language Processing}, pages 106--118, Abu Dhabi, United Arab
  Emirates. Association for Computational Linguistics.

\bibitem[{Chowdhery et~al.(2022)Chowdhery, Narang, Devlin, Bosma, Mishra,
  Roberts, Barham, Chung, Sutton, and Gehrmann~et al}]{chowdhery2022palm}
Aakanksha Chowdhery, Sharan Narang, Jacob Devlin, Maarten Bosma, Gaurav Mishra,
  Adam Roberts, Paul Barham, Hyung~Won Chung, Charles Sutton, and Sebastian
  Gehrmann~et al. 2022.
\newblock {PaLM}: Scaling language modeling with pathways.
\newblock \emph{arXiv preprint arXiv:2204.02311}.

\bibitem[{Clifton et~al.(2020)Clifton, Reddy, Yu, Pappu, Rezapour, Bonab,
  Eskevich, Jones, Karlgren, Carterette, and Jones}]{clifton-etal-2020-100000}
Ann Clifton, Sravana Reddy, Yongze Yu, Aasish Pappu, Rezvaneh Rezapour, Hamed
  Bonab, Maria Eskevich, Gareth Jones, Jussi Karlgren, Ben Carterette, and
  Rosie Jones. 2020.
\newblock \href {https://doi.org/10.18653/v1/2020.coling-main.519} {100,000
  podcasts: A spoken {E}nglish document corpus}.
\newblock In \emph{Proceedings of the 28th International Conference on
  Computational Linguistics}, pages 5903--5917, Barcelona, Spain (Online).
  International Committee on Computational Linguistics.

\bibitem[{Deutsch et~al.(2021)Deutsch, Bedrax-Weiss, and
  Roth}]{deutsch-etal-2021-towards}
Daniel Deutsch, Tania Bedrax-Weiss, and Dan Roth. 2021.
\newblock \href {https://doi.org/10.1162/tacl_a_00397} {Towards
  question-answering as an automatic metric for evaluating the content quality
  of a summary}.
\newblock \emph{Transactions of the Association for Computational Linguistics},
  9:774--789.

\bibitem[{Erkan and Radev(2004)}]{Erkan:2004}
G\"{u}nes Erkan and Dragomir~R. Radev. 2004.
\newblock \href {https://www.aaai.org/Papers/JAIR/Vol22/JAIR-2214.pdf}
  {{LexRank}: {G}raph-based lexical centrality as salience in text
  summarization}.
\newblock \emph{Journal of Artificial Intelligence Research}.

\bibitem[{Fabbri et~al.(2019)Fabbri, Li, She, Li, and
  Radev}]{fabbri-etal-2019-multi}
Alexander Fabbri, Irene Li, Tianwei She, Suyi Li, and Dragomir Radev. 2019.
\newblock \href {https://doi.org/10.18653/v1/P19-1102} {Multi-news: A
  large-scale multi-document summarization dataset and abstractive hierarchical
  model}.
\newblock In \emph{Proceedings of the 57th Annual Meeting of the Association
  for Computational Linguistics}, pages 1074--1084, Florence, Italy.
  Association for Computational Linguistics.

\bibitem[{Fabbri et~al.(2021)Fabbri, Kry{\'s}ci{\'n}ski, McCann, Xiong, Socher,
  and Radev}]{Fabbri:2021}
Alexander~R. Fabbri, Wojciech Kry{\'s}ci{\'n}ski, Bryan McCann, Caiming Xiong,
  Richard Socher, and Dragomir Radev. 2021.
\newblock \href {https://doi.org/10.1162/tacl_a_00373} {{SummEval:
  Re-evaluating Summarization Evaluation}}.
\newblock \emph{Transactions of the Association for Computational Linguistics},
  9:391--409.

\bibitem[{Flynn(2022)}]{Flynn:2022}
Jack Flynn. 2022.
\newblock 28 incredible meeting statistics: Virtual, zoom, in-person meetings
  and productivity.
\newblock \emph{https://www.zippia.com/advice/meeting-statistics/}.

\bibitem[{Gao and Wan(2022)}]{gao-wan-2022-dialsummeval}
Mingqi Gao and Xiaojun Wan. 2022.
\newblock \href {https://doi.org/10.18653/v1/2022.naacl-main.418}
  {{D}ial{S}umm{E}val: Revisiting summarization evaluation for dialogues}.
\newblock In \emph{Proceedings of the 2022 Conference of the North American
  Chapter of the Association for Computational Linguistics: Human Language
  Technologies}, pages 5693--5709, Seattle, United States. Association for
  Computational Linguistics.

\bibitem[{Gliwa et~al.(2019)Gliwa, Mochol, Biesek, and
  Wawer}]{gliwa-etal-2019-samsum}
Bogdan Gliwa, Iwona Mochol, Maciej Biesek, and Aleksander Wawer. 2019.
\newblock \href {https://doi.org/10.18653/v1/D19-5409} {{SAMS}um corpus: A
  human-annotated dialogue dataset for abstractive summarization}.
\newblock In \emph{Proceedings of the 2nd Workshop on New Frontiers in
  Summarization}, pages 70--79, Hong Kong, China. Association for Computational
  Linguistics.

\bibitem[{Goldsack et~al.(2022)Goldsack, Zhang, Lin, and
  Scarton}]{goldsack2022}
Tomas Goldsack, Zhihao Zhang, Chenghua Lin, and Carolina Scarton. 2022.
\newblock Making science simple: Corpora for the lay summarisation of
  scientific literature.
\newblock \emph{arXiv preprint arxiv:2210.09932}.

\bibitem[{Goyal et~al.(2022)Goyal, Li, and Durrett}]{goyal2022news}
Tanya Goyal, Junyi~Jessy Li, and Greg Durrett. 2022.
\newblock News summarization and evaluation in the era of gpt-3.
\newblock \emph{arXiv preprint arXiv:2209.12356}.

\bibitem[{Grusky et~al.(2018)Grusky, Naaman, and
  Artzi}]{grusky-etal-2018-newsroom}
Max Grusky, Mor Naaman, and Yoav Artzi. 2018.
\newblock \href {https://doi.org/10.18653/v1/N18-1065} {{N}ewsroom: A dataset
  of 1.3 million summaries with diverse extractive strategies}.
\newblock In \emph{Proceedings of the 2018 Conference of the North {A}merican
  Chapter of the Association for Computational Linguistics: Human Language
  Technologies, Volume 1 (Long Papers)}, pages 708--719, New Orleans,
  Louisiana. Association for Computational Linguistics.

\bibitem[{Huang et~al.(2021)Huang, Cao, Parulian, Ji, and
  Wang}]{huang-etal-2021-efficient}
Luyang Huang, Shuyang Cao, Nikolaus Parulian, Heng Ji, and Lu~Wang. 2021.
\newblock \href {https://doi.org/10.18653/v1/2021.naacl-main.112} {Efficient
  attentions for long document summarization}.
\newblock In \emph{Proceedings of the 2021 Conference of the North American
  Chapter of the Association for Computational Linguistics: Human Language
  Technologies}, pages 1419--1436, Online. Association for Computational
  Linguistics.

\bibitem[{{Janin} et~al.(2003){Janin}, {Baron}, {Edwards}, {Ellis}, {Gelbart},
  {Morgan}, {Peskin}, {Pfau}, {Shriberg}, {Stolcke}, and
  {Wooters}}]{Janin:2003}
A.~{Janin}, D.~{Baron}, J.~{Edwards}, D.~{Ellis}, D.~{Gelbart}, N.~{Morgan},
  B.~{Peskin}, T.~{Pfau}, E.~{Shriberg}, A.~{Stolcke}, and C.~{Wooters}. 2003.
\newblock The {ICSI} meeting corpus.
\newblock In \emph{Proceedings of the 2003 IEEE International Conference on
  Acoustics, Speech, and Signal Processing (ICASSP)}.

\bibitem[{Khalman et~al.(2021)Khalman, Zhao, and
  Saleh}]{khalman-etal-2021-forumsum-multi}
Misha Khalman, Yao Zhao, and Mohammad Saleh. 2021.
\newblock \href {https://doi.org/10.18653/v1/2021.findings-emnlp.391}
  {{F}orum{S}um: A multi-speaker conversation summarization dataset}.
\newblock In \emph{Findings of the Association for Computational Linguistics:
  EMNLP 2021}, pages 4592--4599, Punta Cana, Dominican Republic. Association
  for Computational Linguistics.

\bibitem[{Koay et~al.(2020)Koay, Roustai, Dai, Burns, Kerrigan, and
  Liu}]{koay-etal-2020-domain}
Jia~Jin Koay, Alexander Roustai, Xiaojin Dai, Dillon Burns, Alec Kerrigan, and
  Fei Liu. 2020.
\newblock \href {https://doi.org/10.18653/v1/2020.coling-main.499} {How domain
  terminology affects meeting summarization performance}.
\newblock In \emph{Proceedings of the 28th International Conference on
  Computational Linguistics}, pages 5689--5695, Barcelona, Spain (Online).
  International Committee on Computational Linguistics.

\bibitem[{Koay et~al.(2021)Koay, Roustai, Dai, and
  Liu}]{koay-etal-2021-sliding}
Jia~Jin Koay, Alexander Roustai, Xiaojin Dai, and Fei Liu. 2021.
\newblock \href {https://doi.org/10.18653/v1/2021.naacl-srw.10} {A
  sliding-window approach to automatic creation of meeting minutes}.
\newblock In \emph{Proceedings of the 2021 Conference of the North American
  Chapter of the Association for Computational Linguistics: Student Research
  Workshop}, pages 68--75, Online. Association for Computational Linguistics.

\bibitem[{Kornilova and Eidelman(2019)}]{kornilova-eidelman-2019-billsum}
Anastassia Kornilova and Vladimir Eidelman. 2019.
\newblock \href {https://doi.org/10.18653/v1/D19-5406} {{B}ill{S}um: A corpus
  for automatic summarization of {US} legislation}.
\newblock In \emph{Proceedings of the 2nd Workshop on New Frontiers in
  Summarization}, pages 48--56, Hong Kong, China. Association for Computational
  Linguistics.

\bibitem[{Kratochvil et~al.(2020)Kratochvil, Polak, and
  Bojar}]{kratochvil-etal-2020-large}
Jonas Kratochvil, Peter Polak, and Ondrej Bojar. 2020.
\newblock \href {https://aclanthology.org/2020.lrec-1.781} {Large corpus of
  {C}zech parliament plenary hearings}.
\newblock In \emph{Proceedings of the Twelfth Language Resources and Evaluation
  Conference}, pages 6363--6367, Marseille, France. European Language Resources
  Association.

\bibitem[{Laleye et~al.(2020)Laleye, de~Chalendar, Blani{\'e}, Brouquet, and
  Behnamou}]{laleye-etal-2020-french}
Fr{\'e}jus A.~A. Laleye, Ga{\"e}l de~Chalendar, Antonia Blani{\'e}, Antoine
  Brouquet, and Dan Behnamou. 2020.
\newblock \href {https://aclanthology.org/2020.lrec-1.72} {A {F}rench medical
  conversations corpus annotated for a virtual patient dialogue system}.
\newblock In \emph{Proceedings of the Twelfth Language Resources and Evaluation
  Conference}, pages 574--580, Marseille, France. European Language Resources
  Association.

\bibitem[{Lewis et~al.(2020)Lewis, Liu, Goyal, Ghazvininejad, Mohamed, Levy,
  Stoyanov, and Zettlemoyer}]{lewis-etal-2020-bart}
Mike Lewis, Yinhan Liu, Naman Goyal, Marjan Ghazvininejad, Abdelrahman Mohamed,
  Omer Levy, Veselin Stoyanov, and Luke Zettlemoyer. 2020.
\newblock \href {https://doi.org/10.18653/v1/2020.acl-main.703} {{BART}:
  Denoising sequence-to-sequence pre-training for natural language generation,
  translation, and comprehension}.
\newblock In \emph{Proceedings of the 58th Annual Meeting of the Association
  for Computational Linguistics}, pages 7871--7880, Online. Association for
  Computational Linguistics.

\bibitem[{Li et~al.(2019)Li, Zhang, Ji, and Radke}]{li-etal-2019-keep}
Manling Li, Lingyu Zhang, Heng Ji, and Richard~J. Radke. 2019.
\newblock \href {https://doi.org/10.18653/v1/P19-1210} {Keep meeting summaries
  on topic: Abstractive multi-modal meeting summarization}.
\newblock In \emph{Proceedings of the 57th Annual Meeting of the Association
  for Computational Linguistics}, pages 2190--2196, Florence, Italy.
  Association for Computational Linguistics.

\bibitem[{Lin(2004)}]{lin-2004-rouge}
Chin-Yew Lin. 2004.
\newblock \href {https://aclanthology.org/W04-1013} {{ROUGE}: A package for
  automatic evaluation of summaries}.
\newblock In \emph{Text Summarization Branches Out}, pages 74--81, Barcelona,
  Spain. Association for Computational Linguistics.

\bibitem[{Mihalcea and Tarau(2004)}]{mihalcea-tarau-2004-textrank}
Rada Mihalcea and Paul Tarau. 2004.
\newblock \href {https://aclanthology.org/W04-3252} {{T}ext{R}ank: Bringing
  order into text}.
\newblock In \emph{Proceedings of the 2004 Conference on Empirical Methods in
  Natural Language Processing}, pages 404--411, Barcelona, Spain. Association
  for Computational Linguistics.

\bibitem[{Moramarco et~al.(2021)Moramarco, Papadopoulos~Korfiatis, Savkov, and
  Reiter}]{moramarco-etal-2021-preliminary}
Francesco Moramarco, Alex Papadopoulos~Korfiatis, Aleksandar Savkov, and Ehud
  Reiter. 2021.
\newblock \href {https://aclanthology.org/2021.humeval-1.7} {A preliminary
  study on evaluating consultation notes with post-editing}.
\newblock In \emph{Proceedings of the Workshop on Human Evaluation of NLP
  Systems (HumEval)}, pages 62--68, Online. Association for Computational
  Linguistics.

\bibitem[{Murray et~al.(2010)Murray, Carenini, and
  Ng}]{murray-etal-2010-generating}
Gabriel Murray, Giuseppe Carenini, and Raymond Ng. 2010.
\newblock \href {https://aclanthology.org/W10-4211} {Generating and validating
  abstracts of meeting conversations: a user study}.
\newblock In \emph{Proceedings of the 6th International Natural Language
  Generation Conference}. Association for Computational Linguistics.

\bibitem[{Narayan et~al.(2018)Narayan, Cohen, and
  Lapata}]{narayan-etal-2018-dont}
Shashi Narayan, Shay~B. Cohen, and Mirella Lapata. 2018.
\newblock \href {https://doi.org/10.18653/v1/D18-1206} {Don{'}t give me the
  details, just the summary! topic-aware convolutional neural networks for
  extreme summarization}.
\newblock In \emph{Proceedings of the 2018 Conference on Empirical Methods in
  Natural Language Processing}, pages 1797--1807, Brussels, Belgium.
  Association for Computational Linguistics.

\bibitem[{Nedoluzhko et~al.(2022)Nedoluzhko, Singh, Hled{\'\i}kov{\'a}, Ghosal,
  and Bojar}]{nedoluzhko-etal-2022-elitr}
Anna Nedoluzhko, Muskaan Singh, Marie Hled{\'\i}kov{\'a}, Tirthankar Ghosal,
  and Ond{\v{r}}ej Bojar. 2022.
\newblock \href {https://aclanthology.org/2022.lrec-1.340} {{ELITR} minuting
  corpus: A novel dataset for automatic minuting from multi-party meetings in
  {E}nglish and {C}zech}.
\newblock In \emph{Proceedings of the Thirteenth Language Resources and
  Evaluation Conference}, pages 3174--3182, Marseille, France. European
  Language Resources Association.

\bibitem[{Ni et~al.(2021)Ni, Azerbayev, Mutuma, Feng, Zhang, Yu, Awadallah, and
  Radev}]{ni-etal-2021-summertime}
Ansong Ni, Zhangir Azerbayev, Mutethia Mutuma, Troy Feng, Yusen Zhang, Tao Yu,
  Ahmed~Hassan Awadallah, and Dragomir Radev. 2021.
\newblock \href {https://doi.org/10.18653/v1/2021.emnlp-demo.37}
  {{S}ummer{T}ime: Text summarization toolkit for non-experts}.
\newblock In \emph{Proceedings of the 2021 Conference on Empirical Methods in
  Natural Language Processing: System Demonstrations}, pages 329--338, Online
  and Punta Cana, Dominican Republic. Association for Computational
  Linguistics.

\bibitem[{Oya et~al.(2014)Oya, Mehdad, Carenini, and
  Ng}]{oya-etal-2014-template}
Tatsuro Oya, Yashar Mehdad, Giuseppe Carenini, and Raymond Ng. 2014.
\newblock \href {https://doi.org/10.3115/v1/W14-4407} {A template-based
  abstractive meeting summarization: Leveraging summary and source text
  relationships}.
\newblock In \emph{Proceedings of the 8th International Natural Language
  Generation Conference ({INLG})}, pages 45--53, Philadelphia, Pennsylvania,
  U.S.A. Association for Computational Linguistics.

\bibitem[{Papalampidi et~al.(2020)Papalampidi, Keller, Frermann, and
  Lapata}]{papalampidi-etal-2020-screenplay}
Pinelopi Papalampidi, Frank Keller, Lea Frermann, and Mirella Lapata. 2020.
\newblock \href {https://doi.org/10.18653/v1/2020.acl-main.174} {Screenplay
  summarization using latent narrative structure}.
\newblock In \emph{Proceedings of the 58th Annual Meeting of the Association
  for Computational Linguistics}, pages 1920--1933, Online. Association for
  Computational Linguistics.

\bibitem[{Post(2018)}]{post-2018-call}
Matt Post. 2018.
\newblock \href {https://doi.org/10.18653/v1/W18-6319} {A call for clarity in
  reporting {BLEU} scores}.
\newblock In \emph{Proceedings of the Third Conference on Machine Translation:
  Research Papers}, pages 186--191, Brussels, Belgium. Association for
  Computational Linguistics.

\bibitem[{Raffel et~al.(2020)Raffel, Shazeer, Roberts, Lee, Narang, Matena,
  Zhou, Li, and Liu}]{raffel2020exploring}
Colin Raffel, Noam Shazeer, Adam Roberts, Katherine Lee, Sharan Narang, Michael
  Matena, Yanqi Zhou, Wei Li, and Peter~J. Liu. 2020.
\newblock Exploring the limits of transfer learning with a unified text-to-text
  transformer.
\newblock \emph{J. Mach. Learn. Res.}, 21(1).

\bibitem[{Renals et~al.(2007)Renals, Hain, and Bourlard}]{Renals:2007}
Steve Renals, Thomas Hain, and Herve Bourlard. 2007.
\newblock \href {https://doi.org/10.1109/ASRU.2007.4430116} {Recognition and
  understanding of meetings the ami and amida projects}.
\newblock In \emph{2007 IEEE Workshop on Automatic Speech Recognition and
  Understanding (ASRU)}, pages 238--247.

\bibitem[{Shang et~al.(2018)Shang, Ding, Zhang, Tixier, Meladianos,
  Vazirgiannis, and Lorr{\'e}}]{shang-etal-2018-unsupervised}
Guokan Shang, Wensi Ding, Zekun Zhang, Antoine Tixier, Polykarpos Meladianos,
  Michalis Vazirgiannis, and Jean-Pierre Lorr{\'e}. 2018.
\newblock \href {https://doi.org/10.18653/v1/P18-1062} {Unsupervised
  abstractive meeting summarization with multi-sentence compression and
  budgeted submodular maximization}.
\newblock In \emph{Proceedings of the 56th Annual Meeting of the Association
  for Computational Linguistics (Volume 1: Long Papers)}, pages 664--674,
  Melbourne, Australia. Association for Computational Linguistics.

\bibitem[{Song et~al.(2022)Song, Li, Wang, Yu, and
  Liu}]{song-etal-2022-towards}
Kaiqiang Song, Chen Li, Xiaoyang Wang, Dong Yu, and Fei Liu. 2022.
\newblock \href {https://doi.org/10.18653/v1/2022.acl-long.302} {Towards
  abstractive grounded summarization of podcast transcripts}.
\newblock In \emph{Proceedings of the 60th Annual Meeting of the Association
  for Computational Linguistics (Volume 1: Long Papers)}, pages 4407--4418,
  Dublin, Ireland. Association for Computational Linguistics.

\bibitem[{Tardy et~al.(2020)Tardy, Janiszek, Est{\`e}ve, and
  Nguyen}]{tardy-etal-2020-align}
Paul Tardy, David Janiszek, Yannick Est{\`e}ve, and Vincent Nguyen. 2020.
\newblock \href {https://aclanthology.org/2020.lrec-1.829} {Align then
  summarize: Automatic alignment methods for summarization corpus creation}.
\newblock In \emph{Proceedings of the Twelfth Language Resources and Evaluation
  Conference}, pages 6718--6724, Marseille, France. European Language Resources
  Association.

\bibitem[{Wang and Cardie(2013)}]{wang-cardie-2013-domain}
Lu~Wang and Claire Cardie. 2013.
\newblock \href {https://aclanthology.org/P13-1137} {Domain-independent
  abstract generation for focused meeting summarization}.
\newblock In \emph{Proceedings of the 51st Annual Meeting of the Association
  for Computational Linguistics (Volume 1: Long Papers)}, pages 1395--1405,
  Sofia, Bulgaria. Association for Computational Linguistics.

\bibitem[{Wei et~al.(2022)Wei, Tay, Bommasani, Raffel, Zoph, Borgeaud,
  Yogatama, Bosma, Zhou, Metzler, Chi, Hashimoto, Vinyals, Liang, Dean, and
  Fedus}]{wei2022emergent}
Jason Wei, Yi~Tay, Rishi Bommasani, Colin Raffel, Barret Zoph, Sebastian
  Borgeaud, Dani Yogatama, Maarten Bosma, Denny Zhou, Donald Metzler, Ed~H.
  Chi, Tatsunori Hashimoto, Oriol Vinyals, Percy Liang, Jeff Dean, and William
  Fedus. 2022.
\newblock \href {https://openreview.net/forum?id=yzkSU5zdwD} {Emergent
  abilities of large language models}.
\newblock \emph{Transactions on Machine Learning Research}.
\newblock Survey Certification.

\bibitem[{Zechner(2002)}]{zechner-2002-automatic}
Klaus Zechner. 2002.
\newblock \href {https://doi.org/10.1162/089120102762671945} {Automatic
  summarization of open-domain multiparty dialogues in diverse genres}.
\newblock \emph{Computational Linguistics}, 28(4):447--485.

\bibitem[{Zeng et~al.(2020)Zeng, Yang, Ju, Yang, Wang, Zhang, Zhou, Zeng, Dong,
  Zhang, Fang, Zhu, Chen, and Xie}]{zeng-etal-2020-meddialog}
Guangtao Zeng, Wenmian Yang, Zeqian Ju, Yue Yang, Sicheng Wang, Ruisi Zhang,
  Meng Zhou, Jiaqi Zeng, Xiangyu Dong, Ruoyu Zhang, Hongchao Fang, Penghui Zhu,
  Shu Chen, and Pengtao Xie. 2020.
\newblock \href {https://doi.org/10.18653/v1/2020.emnlp-main.743}
  {{M}ed{D}ialog: Large-scale medical dialogue datasets}.
\newblock In \emph{Proceedings of the 2020 Conference on Empirical Methods in
  Natural Language Processing (EMNLP)}, pages 9241--9250, Online. Association
  for Computational Linguistics.

\bibitem[{Zhang et~al.(2020{\natexlab{a}})Zhang, Zhao, Saleh, and
  Liu}]{zhang2020pegasus}
Jingqing Zhang, Yao Zhao, Mohammad Saleh, and Peter Liu. 2020{\natexlab{a}}.
\newblock Pegasus: Pre-training with extracted gap-sentences for abstractive
  summarization.
\newblock In \emph{International Conference on Machine Learning}, pages
  11328--11339. PMLR.

\bibitem[{Zhang et~al.(2020{\natexlab{b}})Zhang, Kishore, Wu, Weinberger, and
  Artzi}]{Zhang:2020:BERTScore}
Tianyi Zhang, Varsha Kishore, Felix Wu, Kilian~Q. Weinberger, and Yoav Artzi.
  2020{\natexlab{b}}.
\newblock {BERTScore}: Evaluating text generation with {BERT}.
\newblock In \emph{International Conference on Learning Representations}.

\bibitem[{Zhang et~al.(2023)Zhang, Ladhak, Durmus, Liang, McKeown, and
  Hashimoto}]{zhang2023benchmarking}
Tianyi Zhang, Faisal Ladhak, Esin Durmus, Percy Liang, Kathleen McKeown, and
  Tatsunori~B. Hashimoto. 2023.
\newblock Benchmarking large language models for news summarization.
\newblock \emph{arXiv preprint arxiv:2301.13848}.

\bibitem[{Zhang et~al.(2022)Zhang, Ni, Mao, Wu, Zhu, Deb, Awadallah, Radev, and
  Zhang}]{zhang-etal-2022-summn}
Yusen Zhang, Ansong Ni, Ziming Mao, Chen~Henry Wu, Chenguang Zhu, Budhaditya
  Deb, Ahmed Awadallah, Dragomir Radev, and Rui Zhang. 2022.
\newblock \href {https://doi.org/10.18653/v1/2022.acl-long.112} {{S}umm$^n$: A
  multi-stage summarization framework for long input dialogues and documents}.
\newblock In \emph{Proceedings of the 60th Annual Meeting of the Association
  for Computational Linguistics (Volume 1: Long Papers)}, pages 1592--1604,
  Dublin, Ireland. Association for Computational Linguistics.

\bibitem[{Zhao et~al.(2019)Zhao, Peyrard, Liu, Gao, Meyer, and
  Eger}]{zhao-etal-2019-moverscore}
Wei Zhao, Maxime Peyrard, Fei Liu, Yang Gao, Christian~M. Meyer, and Steffen
  Eger. 2019.
\newblock \href {https://doi.org/10.18653/v1/D19-1053} {{M}over{S}core: Text
  generation evaluating with contextualized embeddings and earth mover
  distance}.
\newblock In \emph{Proceedings of the 2019 Conference on Empirical Methods in
  Natural Language Processing and the 9th International Joint Conference on
  Natural Language Processing (EMNLP-IJCNLP)}, pages 563--578, Hong Kong,
  China. Association for Computational Linguistics.

\bibitem[{Zhong et~al.(2022)Zhong, Liu, Xu, Zhu, and Zeng}]{zhong2022dialoglm}
Ming Zhong, Yang Liu, Yichong Xu, Chenguang Zhu, and Michael Zeng. 2022.
\newblock Dialoglm: Pre-trained model for long dialogue understanding and
  summarization.
\newblock In \emph{Proceedings of the 36th AAAI Conference on Artificial
  Intelligence}.

\bibitem[{Zhong et~al.(2021)Zhong, Yin, Yu, Zaidi, Mutuma, Jha, Awadallah,
  Celikyilmaz, Liu, Qiu, and Radev}]{zhong2021qmsum}
Ming Zhong, Da~Yin, Tao Yu, Ahmad Zaidi, Mutethia Mutuma, Rahul Jha,
  Ahmed~Hassan Awadallah, Asli Celikyilmaz, Yang Liu, Xipeng Qiu, and Dragomir
  Radev. 2021.
\newblock Qmsum: A new benchmark for query-based multi-domain meeting
  summarization.
\newblock \emph{arXiv preprint arxiv:2104.05938}.

\bibitem[{Zhu et~al.(2021)Zhu, Liu, Mei, and Zeng}]{zhu-etal-2021-mediasum}
Chenguang Zhu, Yang Liu, Jie Mei, and Michael Zeng. 2021.
\newblock \href {https://doi.org/10.18653/v1/2021.naacl-main.474}
  {{M}edia{S}um: A large-scale media interview dataset for dialogue
  summarization}.
\newblock In \emph{Proceedings of the 2021 Conference of the North American
  Chapter of the Association for Computational Linguistics: Human Language
  Technologies}, pages 5927--5934, Online. Association for Computational
  Linguistics.

\bibitem[{Zhu et~al.(2020)Zhu, Xu, Zeng, and
  Huang}]{zhu-etal-2020-hierarchical}
Chenguang Zhu, Ruochen Xu, Michael Zeng, and Xuedong Huang. 2020.
\newblock \href {https://doi.org/10.18653/v1/2020.findings-emnlp.19} {A
  hierarchical network for abstractive meeting summarization with cross-domain
  pretraining}.
\newblock In \emph{Findings of the Association for Computational Linguistics:
  EMNLP 2020}, pages 194--203, Online. Association for Computational
  Linguistics.

\end{thebibliography}
\bibliographystyle{acl_natbib}

\appendix

\section{Experimental Settings}
\label{sec:appendix}

Our implementation details and hyperparameter settings for both extractive systems and abstractive systems with fine-tuning are shown in Table~\ref{tab:params}. 
We use \texttt{text-davinci-003} version of GPT-3 in our experiments. We follow the convention of Goyal et al.~\shortcite{goyal2022news} and use the following prompt asking the model to summarize a transcript in two sentences:
\vspace{0.07in}
\noindent\emph{{Article:\{\{article\}\}}}\\
\noindent\emph{{Summarize the above article in N sentences.}}

\section{Comparison to ICSI/AMI}
\label{sec:additional}

By introducing a corpus, we aim to spur research and development in the area of meeting summarization.  However, meetings often pertain to specialized domains and exhibit unique structures. Our preliminary experiments suggest that a BART summarizer fine-tuned for our dataset does not perform optimally on the ICSI/AMI datasets. In particular, the ICSI meetings pose a challenge as they are research seminars conducted by a group of speech researchers, whereas our dataset is collected from city councils in the U.S.

\begin{table}[t]
\setlength{\tabcolsep}{5pt}
\renewcommand{\arraystretch}{1.18}
\centering
\begin{fontpbk}
\begin{small}
\begin{tabular}{|l|}
\hline
\textbf{\texttt{Extractive Oracle}}\\
\hdashline
We use the implementation provided by Paul Tardy:\\
\url{github.com/pltrdy/extoracle_summarization}\\
(a) ``-length\_oracle'' sets the output to have the same\\
number of sentences as the reference summary.\\
(b) ``-method greedy -length 999'' allows the greedy \\
algorithm to select an optimal number of sentences\\
that yield the highest (R1+R2) scores.\\
In this paper, we report results using option (b).\\
\hline
\hline
\textbf{\texttt{LexRank}} and \textbf{\texttt{TextRank}}\\
\hdashline
We use SummerTime's implementation of LexRank\\
 and TextRank with default parameters.\\
\url{https://github.com/Yale-LILY/SummerTime}\\
For each meeting segment, 2 sentences are extracted.\\
\hline
\hline
\textbf{\texttt{BART}}\\
\hdashline
The BART model is initialized using \href{https://huggingface.co/facebook/bart-large-cnn}{bart-large-cnn}:\\
The default model parameters are used, \\
with some of the important ones listed below.\\
- max input sequence length: 1,024 tokens\\
- min output length: 56 tokens\\
- max output length: 142 tokens\\
- beam width: 4\\
- length penalty: 2.0\\
- initial learning rate: 2.5e-6\\
\hline
\hline
\textbf{\texttt{Pegasus}}\\
\hdashline
Pegasus is initialized using \href{https://huggingface.co/google/pegasus-xsum}{google/pegasus-xsum}\\
- max sequence length: 512\\
- max output length: 64\\
- beam width: 8\\
- length penalty: 0.6\\
- initial learning rate: 2.5e-6\\
\hline
\hline
\textbf{\texttt{Longformer}}\\
\hdashline
Longformer is initialized using patrickvonplaten/\\
\href{https://huggingface.co/patrickvonplaten/longformer2roberta-cnn_dailymail-fp16}{longformer2roberta-cnn\_dailymail-fp16}\\
- max sequence length: 4,098\\
- min output length: 56\\
- max output length: 142\\
- beam width: 1\\
- length penalty: 1.0\\
- Initial Learning Rate: 2.5e-6\\
\hline
\hline
\textbf{\texttt{HMNet}}\\
\hdashline
We use the implementation of Zhu et al.~\shortcite{zhu-etal-2020-hierarchical}.\\
HMNet is initialized using \href{https://github.com/microsoft/HMNet/tree/main/ExampleInitModel/HMNet-pretrained}{HMNet-pretrained}\\
- max sequence length: 8300\\
- min output length: 10\\
- max output length: 300\\
- beam width: 6\\
- initial learning rate: 1e-4\\
\hline
\hline
\textbf{\texttt{DialogLM}}\\
\hdashline
We use the original \href{https://github.com/microsoft/DialogLM}{DialogLM source implementation}.\\
- max sequence length: 5,632\\
- min output length: 10\\
- max output length: 300\\
- beam width: 6\\
- Initial Learning Rate: 7e-5\\
\hline
\end{tabular}
\end{small}
\end{fontpbk}
\caption{Implementation details and hyperparameter settings for extractive systems and abstractive systems. 
}
\label{tab:params}
\vspace{-0.1in}
\end{table}

\end{document}